\documentclass[runningheads]{llncs}
\usepackage{graphicx}
\usepackage{tikz}
\usepackage{comment}
\usepackage{amsmath,amssymb} % define this before the line numbering.
\usepackage{color}
\usepackage{xcolor}
\usepackage{arydshln}

\usepackage[accsupp]{axessibility}  % Improves PDF readability for those with disabilities.

\begin{document}
% \linenumbers
\pagestyle{headings}
\mainmatter
\def\ECCVSubNumber{3}  % Insert your submission number here

\title{Comparison of semi-supervised learning methods for High Content Screening quality control} % Replace with your title

\begin{comment}
\titlerunning{ECCV-22 submission ID \ECCVSubNumber} 
\authorrunning{ECCV-22 submission ID \ECCVSubNumber} 
\author{Anonymous ECCV submission}
\institute{Paper ID \ECCVSubNumber}
\end{comment}
%******************

%\begin{comment}
\titlerunning{HCS Quality Control: transfer or self-supervise?}

\author{Umar Masud\inst{4,*}  \and
Ethan Cohen\inst{1,2,*}  \and
Ihab Bendidi \inst{1,3} \and
Guillaume Bollot \inst{2} \and
Auguste Genovesio \inst{1} 
}

\renewcommand{\thefootnote}{\fnsymbol{footnote}}
\footnotetext[1]{Equal co-contribution.}
\authorrunning{U.Masud, E.Cohen et al.}
\institute{IBENS, Ecole Normale Sup\'erieure, Paris, France \and
SYNSIGHT, Evry, France \and Minos Biosciences, Paris, France \and
Jamia Millia Islamia, New Delhi, India\\
\email{ecohen@bio.ens.psl.eu, auguste.genovesio@ens.psl.edu}}

%\end{comment}
%******************
\maketitle

\begin{abstract}
Progress in automated microscopy and quantitative image analysis has promoted high-content screening (HCS) as an efficient drug discovery and research tool. While HCS offers to quantify complex cellular phenotypes from images at high throughput, this process can be obstructed by image aberrations such as out-of-focus image blur, fluorophore saturation, debris, a high level of noise, unexpected auto-fluorescence or empty images. While this issue has received moderate attention in the literature, overlooking these artefacts can seriously hamper downstream image processing tasks and hinder detection of subtle phenotypes. It is therefore of primary concern, and a prerequisite, to use quality control in HCS. In this work, we evaluate deep learning options that do not require extensive image annotations to provide a straightforward and easy to use semi-supervised learning solution to this issue. Concretely, we compared the efficacy of recent self-supervised and transfer learning approaches to provide a base encoder to a high throughput artefact image detector. The results of this study suggest that transfer learning methods should be preferred for this task as they not only performed best here but present the advantage of not requiring sensitive hyperparameter settings nor extensive additional training.

\keywords{Cell-based assays  \and  Image analysis \and Deep learning \and Self-supervised learning}
\end{abstract}

\section{Introduction}
Image analysis solutions are heavily used in microscopy. They enable the extraction of quantitative information from cells, tissues and organisms. These methods and tools have proven to be especially useful for high-content screening (HCS), an automated approach that produces a large amount of microscopy image data, to study various mechanisms and identify genetic and chemical modulators in drug discovery and research~\cite{hcs}. However, the success of an HCS screen is often related to the dataset quality obtained at end. In practice, abnormalities in image quality are numerous and can lead to imprecise results at best, and erroneous results or false conclusions at worst. Common abnormalities include noise, out-of-focus, presence of debris, blur or image saturation. Furthermore, in some cases, it can also be convenient to exclude images full of dead or floating cells. More importantly, in HCS, manual inspection of all images in a dataset is intractable, as one such screen typically encompasses hundreds of thousands of images. 

Quality control (QC) methods have been investigated for this purpose. Interesting software such as CellProfiler~\cite{cellprofiler} allows end-to-end analysis pipeline with an integrated QC modules. Although powerful, the image quality measures are mainly handcrafted with different computed metrics as described in~\cite{cellsoftware} and therefore hard to generalize. More recently, Yang et al proposed a method to assess microscope image focus using deep learning~\cite{yang}. However, this approach is restricted to a specific type of aberration and does not generalize well to other kinds of artefacts. Besides, learning all types of aberrations from scratch in a supervised manner is hardly tractable, given the diversity of both normal and abnormal image types. It would require systematic annotation of all types of aberrations on each new high-throughput assay, and thus would be utterly time-consuming and hardly feasible in practice. For this task, we thus typically seek a semi-supervised solution that would require annotation of a limited amount of data per assay. 

Transfer learning typically offers such a solution that relies on little supervision \cite{Kensert2019-ve}. A network pretrained on a large annotated image set can be reused directly or fine-tuned with a limited set of annotated images to solve a specific task in another domain. Furthermore, recent breakthroughs in self-supervised learning (SSL), which aim to learn representations without any labels data call for new methods  ~\cite{survey,byol,barlow,vicreg,deepcluster,scan,swav,simclr}. For instance, such a framework was successfully used by Perakis et al \cite{moa} to learn single-cell representations for classification of treatments into mechanisms of action. It was shown that SSL performed better than the more established transfer learning (TL) in several applications. However, it is not a strict rule and not systematically the case as assessed by a recent survey \cite{yang2020}. It is still unclear which approach works better on what type of data and tasks.

In this work, we propose to address this question  in the context of HCS quality control. To this end we performed a comparative study of a range of SSL and TL approaches to detect abnormal single-cell images in a high-content screening dataset with a low amount of annotated assay specific image data. The paper is organized as follows. In Section~\ref{sec:related work}, we briefly describe the various methods we use for transfer and self-supervised representation learning. In Section~\ref{sec:methodology}, we then detail the setup of this comparative study. We then provide experimental results in Section~\ref{sec:results}, and concluding remarks in Section~\ref{sec:concl}.

\section{Related work}
\label{sec:related work}

We seek a method that would provide a robust base encoder to a quality control downstream task where a low amount of annotated data is available. We thought of several options that could be grouped in two categories: transfer learning and self-supervised learning methods.

\subsection{Transfer learning}

Training a deep learning model efficiently necessitates a significant amount of data. In the case of supervised training, it is required that data be annotated with class labels. Transfer learning has become popular to circumvent this issue. It consists in pretraining a network on a large set of annotated images in a given domain, typically a domain where image could be annotated. A variety of tasks in various other domains can then be addressed with decent performance simply by reusing the pretrained network as is or by fine tuning its training on a small available dataset on a specific task in the domain of interest.

In this work, we included three popular networks pretrained with ImageNet for transfer learning. First we used VGG16, a model introduced in 2014 that made a significant improvement over the early AlexNet introduced in 2012, by widening the size of convolutional layer kernels ~\cite{vgg}. We also used ResNet18, a network introduced in 2016 that implements residual connections to make possible the stable training of deeper networks~\cite{resnet}. Finally we used ConvNext, one of the most recent convolutional networks introduced in 2022 that competes favorably with most models, including vision transformer, while maintaining the simplicity and efficiency of ConvNets~\cite{convnetx}. 

\subsection{Self-supervised learning}
In recent years, self-supervised representation learning has gained popularity thanks to its ability to avoid the need for human annotations. It has provided ways to learn useful and robust representation without labeling any data. Most of these approaches rely on a common and simple principle. Two or more random transformations are applied to the same images to produce a set of images containing different views of the same information content. These images are then passed through an encoder that is trained to somewhat encourage learning of a close and invariant representation through the optimization of a given loss function. The loss function varies depending on the method, but once a self-supervised representation is learned, it can be used to solve downstream tasks that may require little to no annotated data. Various kind of SSL mechanisms have been developed, but a wide range of approaches can be summarized in three classes of methods our study encompasses here, namely contrastive, non contrastive and clustering-based methods: 
\begin{enumerate}
\item \textbf{Contrastive learning methods} aim to group similar samples closer and diverse samples farther from one another. Although powerful, such methods still need to find some negative examples via a memory bank or to use a large batch size for end to end learning~\cite{survey}.
\item \textbf{Non-contrastive learning methods} use only positive sample pairs compared to contrastive methods. These approaches proved to learn good representations without the need for a large batch size or memory bank~\cite{byol,barlow,vicreg}.
\item \textbf{Clustering-based learning methods}  ensure that similar samples cluster together but use a clustering algorithm instead of similarity metrics to better generalize by avoiding direct comparison ~\cite{deepcluster,scan,swav}.
\end{enumerate}
In this work, we used a list of methods from the three categories listed above, namely SimCLR~\cite{simclr} for contrastive learning (based on similarity maximisation objective), Barlow Twins~\cite{barlow} and VICReg~\cite{vicreg} for non-contrastive techniques (based on redundancy reduction objective) and DeepCluster~\cite{deepcluster} and SwAV~\cite{swav} for clustering-based methods.

\section{Method}
\label{sec:methodology}

We performed a comparative study that aimed at identifying which of the previously described approaches could be best suited to provide a base encoder, in order to build a classifier for abnormal images from a small annotated image set. In this section we describe the data, the way we perform training for the encoders and the downstream tasks we designed to evaluate and compare them. 

\subsection{Data}
\begin{figure}[h]
\includegraphics[width=0.7\linewidth]{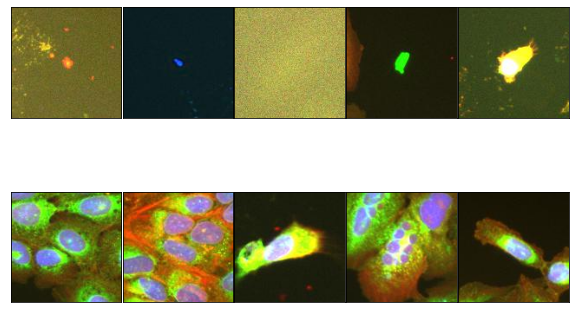}
\centering
\caption{BBBC022 dataset: the first row displays a few abnormal images, the second row shows a few regular images of cells. U2OS cells with Hoechst 33342 staining for nuclei (blue), WGA + phalloidin staining for actin filaments (Red) and MitoTracker staining for mitochondria (green). } \label{fig:samples}
\end{figure}

We used the BBBC022 image set, available from the Broad Bioimage Benchmark Collection to conduct our experiments~\cite{BBBC022,BBBC}. To obtain images at a single-cell level, we cropped a fixed $128 \times 128$ pixel square around the center of each nucleus, resulting in a total of $2,122,341 (\approx2.1M)$ images~\cite{preprocess} . Most of these images were used to train the base encoder when needed (i.e. for SSL methods). Separately, we manually annotated $240$ abnormal and $240$ normal images and split them in a balanced way into training (350) and test (130) sets, with a 50\% ratio of normal images in both the training set and the test set, for the downstream tasks. Some annotated images are displayed in Figure \ref{fig:samples}. Furthermore, we also used $200$ annotated images from the  BBBC021 image set, available from the Broad Bioimage Benchmark Collection to test the generalization of our approach~\cite{BBBC021,BBBC}.
This dataset differs from the BBBC022 in the types of cells used, with U20S cells for BBBC022 and MCF7 cells  for BBBC021 The difference in cell line results in differences in gene expression which translate in different visual features, even with a similar image acquisition process \cite{Yao2019}.

\subsection{Encoder training}
For all TL methods, we used a model pre-trained on ImageNet as an encoder. For SSL methods, we used two networks - first a ResNet18 as encoder, and a fully connected layers (FC layers) as projector. The encoder takes an image as input and outputs a $512$ dimensional vector, which goes as input for the projector network, that is in turn made of $2048$ dimension FC layers and a temperature of $0.07$ for the unsupervised loss. We forward pass the batch of images after producing two different views of them using augmentations. The following augmentations were randomly performed: $90$ degree rotation, flip, transpose, shift and scale. We carefully chose these augmentations so as to keep the trained features relevant to our downstream classification tasks. We use a line search strategy to optimize the hyperparameters of the models. The encoder was afterward trained for $5$ epochs (about 10 million images) for all the SSL methods, with a batch size of $128$, using the SGD optimizer with an initial learning rate of $0.001$ and a momentum of $0.9$. We also used a warm-up cosine scheduler with warm-up epochs set to $1$. For DeepCluster specifically, we set $500$ prototypes and 10 K-means iterations. As for SwAV, we also set the number of prototypes to $500$ and chose a queue size of $2560$. Other projector network parameters were the same as those used in the original papers. We trained models and ran experiments using one Tesla P$100$ GPU with $16$GB vram. All experiments involving SSL methods were done using the solo-learn library~\cite{solo-learn}. Once the encoder was trained, the projector network was discarded and only the ResNet18 network was used for downstream tasks.

\subsection{Downstream Classification tasks}
After training an encoder with each previously described TL or SSL method, we used them to train and test the three following downstream classification tasks, only with the small annotated dataset previously described, after having performed a line search strategy for hyperparameter optimization:
\begin{enumerate}
\item \textbf{K-nn on a frozen encoder output.} We first aimed to evaluate a simple classification setting that did not necessitate any additional training. To this end, we performed a K-Nearest Neighbour (KNN) classification (here we chose k=5) on the $512$ feature vectors output of the encoder.

\item \textbf{Linear classifier trained on a frozen encoder output.} We then evaluated the supervised training of a single dense layer with $2$ output classes on top of the pre-trained encoder. In this setting, the weights of the pre-trained encoder were frozen. We trained this layer for $150$ epochs and with a batch size of $32$. The optimizer was SGD with a learning rate set to $0.001$ and momentum to $0.9$. We also used a step scheduler with gamma value set to $0.1$ at $40$, $70$ and $120$ epochs.

\item \textbf{Linear classifier with fine tuned training of the encoder.} We then used a dense linear layer with $2$ output classes as in the previous settings. However, this time we did not freeze the encoder network and allowed it to pursue training. We trained the models for $50$ epochs. The learning rate was $0.001$ with a momentum of $0.9$ with a SGD optimizer. We also used a step scheduler with a gamma value of $0.1$ at 25 on 40 epochs.
\end{enumerate}

\subsection{Evaluation Criteria} 
We used Accuracy, F$1$-Score and the Area Under Curve (AUC) score to assess the classification results. All displayed values are weighted average for the $2$ classes. The most important metric is the F$1$ score because it takes both false positives and false negatives into account. Thus, the higher the F$1$ score, the better the result. All values mentioned are in percentages.

\section{Results}
\label{sec:results}

\subsection{Evaluations on downstream tasks} 
The results for the Linear Layer classifier and KNN are displayed in Table \ref{tab:linear and  knn}. Among the self-supervised method, we can observe that DeepCluster performs best in both settings and reaches a maximum of $94.57$\% accuracy. SimCLR also performs best with KNN while VICReg performs poorly, dropping to $76.30$\% accuracy. However, none of the SSL methods outperforms the three TL encoders with ConvNext culminating at $98.47$\% accuracy with a Linear Layer classifier. 

Furthermore, the results obtained with fine tune trainings of all the encoder weights are displayed in the first 3 columns of Table \ref{tab:semi supervise}. We can see that with 350 training images (the full annotated training image set), the best results were again obtained with the three TL methods. However, SSL method performed almost as well in this setting with simCLR reaching the best results among the SSL methods with $98.44$\% accuracy.

\begin{table}
  \centering
  \caption{Classification with KNN or a single Linear Layer with a frozen encoder using a 350-image training set. 130 images were used for test.}
  \label{tab:linear and knn}
  \setlength{\tabcolsep}{10pt} % Default value: 6pt
  \renewcommand{\arraystretch}{1.3}
  \begin{tabular}{cccc|ccc}
    \hline
    &\multicolumn{3}{c}{\textbf{KNN}}&\multicolumn{3}{c}{\textbf{Linear Layer}}\\
    \textbf{Method}      &Acc&F1&AUC &Acc&F1&AUC\\
    \hline

    VGG16     &96.09  & 95.94 & 95.38 & 98.24 & 98.12  & 98.09  \\
   
    ResNet18     &\textbf{97.66}  & \textbf{97.51} & \textbf{97.63} & 98.44 & 98.21  & \textbf{98.26}  \\
    
    ConvNext     &97.65  & 97.42 & 97.56 & \textbf{98.47} & \textbf{98.24}  & 98.17  \\
    
\hdashline
    
    SimCLR     &90.62  & 89.63  & 90.30& 91.40 & 90.56  & 91.20   \\
    
    Barlow Twins    &89.84  &88.02  &88.24  & 87.28 & 86.42  & 87.35 \\
    
    VICReg   &76.30  &75.28  &75.13  & 87.76 & 87.36  & 87.40 \\
    
    DeepClusterV2     &90.62  &89.50  &89.81 & 94.57 & 94.02  & 94.14 \\
    
    SwAV   &89.84  &88.52  &89.46 & 94.53 & 94.00  & 94.08     \\
    \hline
  \end{tabular}
\end{table}

\subsection{Effect of a decreasing amount of annotated data}

We also performed an ablation study where the number of training images was gradually decreased. We performed training of  the third task with $350$, $100$, $50$, $25$, and finally just $10$ images. The purpose for decreasing the amount of training images was to evaluate how much supervision the network needs to perform properly. 

The results are displayed in Table \ref{tab:semi supervise}. As the number of training images decrease, VICReg, DeepCluster and SwAV display a drop in performance. With only 25 images, Barlow Twins still produces fairly good results with $94.53$\% accuracy. With just 10 images, the best result among the SSL methods is simCLR with $84.89$\% accuracy. Overall, semi-supervised training can yield good results even with a few images. However, here again, none of the SSL methods outperforms the transfer learning baselines. 

\begin{figure}[h]
\includegraphics[width=0.7\linewidth]{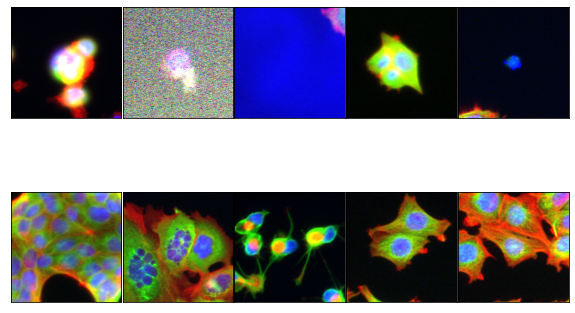}
\centering
\caption{BBBC021 dataset: the first row displays a few abnormal images, the second row displays a few regular cell images. Fixed human MCF7 cells labeled for DNA (blue), actin (red) and B-tubulin (green)} \label{fig:samples_021}
\end{figure}

\subsection{Effect of a domain shift}

To evaluate how these encoders pretrained on BBBC022 or ImageNet could generalize to a different dataset, we tested them on data taken from BBBC021. For this purpose, we considered our best model, the Linear Layer approach with fine tune training of all the encoder weights on $350$ images from the BBBC022 dataset. We then tested it on unseen data taken from BBBC021. We annotated 100 normal and 100 abnormal images from this last dataset for this purpose. Some sample images are displayed in Figure \ref{fig:samples_021}. We made sure to include diverse images in order to thoroughly check the robustness of our trained models.

The results are displayed in Table \ref{tab:generalisation}. Among SSL methods, SimCLR and DeepCluster performed best with respectively $73.66$\% and $72.32$\% accuracy. These results show that some self-supervised learning methods such as simCLR or DeepCluster trained on a large dataset produce features that could generalize a quality control task to an unseen dataset to some extent. However, in accordance with what was observed in previous sections on BBBC022, none of these approaches outperformed the results obtained with the TL encoders.

\begin{table}[]
\centering
\caption{Effect of a decreasing amount of training images on a Linear Layer classifier with a non-frozen encoder. 130 images were used for test.}
\label{tab:semi supervise}
\setlength{\tabcolsep}{5pt} % Default value: 6pt
  \renewcommand{\arraystretch}{1.5}
  \resizebox{\textwidth}{!}{%
\begin{tabular}{cccccccccccccccc}
\hline
\multicolumn{16}{c}{\textbf{Number of Training Images}} \\ \hline
 &
  \multicolumn{3}{c}{\textbf{350}} &
  \multicolumn{3}{c}{\textbf{100}} &
  \multicolumn{3}{c}{\textbf{50}} &
  \multicolumn{3}{c}{\textbf{25}} &
  \multicolumn{3}{c}{\textbf{10}} \\
\textbf{Method} &
  Acc &
  F1 &
  \multicolumn{1}{c|}{AUC} &
  Acc &
  F1 &
  \multicolumn{1}{c|}{AUC} &
  Acc &
  F1 &
  \multicolumn{1}{c|}{AUC} &
  Acc &
  F1 &
  \multicolumn{1}{c|}{AUC} &
  Acc &
  F1 &
  AUC \\ \hline

VGG16 &
  99.08 &
  98.92 &
  \multicolumn{1}{c|}{99.01} &
  99.12 &
  98.83 &
  \multicolumn{1}{c|}{98.89} &
  \textbf{99.22} &
  \textbf{99.18} &
  \multicolumn{1}{c|}{\textbf{99.24}} &
  \textbf{98.44} &
  \textbf{98.19} &
  \multicolumn{1}{c|}{\textbf{98.26}} &
  96.09 &
  95.50 &
  95.81 \\  
ResNet18 &
  99.19 &
  99.11 &
  \multicolumn{1}{c|}{99.02} &
  99.08 &
  98.96 &
  \multicolumn{1}{c|}{99.07} &
  98.54 &
  98.48 &
  \multicolumn{1}{c|}{98.47} &
  97.66 &
  97.81 &
  \multicolumn{1}{c|}{97.62} &
  93.75 &
  93.83 &
  93.54 \\

ConvNext &
  \textbf{99.39} &
  \textbf{99.21} &
  \multicolumn{1}{c|}{\textbf{99.25}} &
  \textbf{99.28} &
  \textbf{99.17} &
  \multicolumn{1}{c|}{\textbf{99.24}} &
  97.66 &
  97.27 &
  \multicolumn{1}{c|}{97.53} &
  97.62 &
  97.49 &
  \multicolumn{1}{c|}{98.02} &
  \textbf{96.87} &
  \textbf{96.68} &
  \textbf{96.55} \\

\hdashline
 
SimCLR &
  98.44 &
  98.21 &
  \multicolumn{1}{c|}{98.34} &
  91.93 &
  92.52 &
  \multicolumn{1}{c|}{93.00} &
  92.97 &
  93.13 &
  \multicolumn{1}{c|}{93.15} &
  92.97 &
  93.17 &
  \multicolumn{1}{c|}{92.80} &
  84.89 &
  84.61 &
  84.32 \\
Barlow Twins &
  98.44 &
  98.14 &
  \multicolumn{1}{c|}{98.12} &
  95.05 &
  95.08 &
  \multicolumn{1}{c|}{94.95} &
  94.53 &
  94.21 &
  \multicolumn{1}{c|}{93.86} &
  94.53 &
  94.12 &
  \multicolumn{1}{c|}{93.65} &
  75.00 &
  80.39 &
  83.14 \\
VICReg &
 98.24 &
  98.00 &
  \multicolumn{1}{c|}{97.88} &
  89.32 &
  89.03 &
  \multicolumn{1}{c|}{89.20} &
  71.01 &
  74.33 &
  \multicolumn{1}{c|}{74.15} &
  72.13 &
  77.37 &
  \multicolumn{1}{c|}{80.22} &
  66.40 &
  72.12 &
  72.00 \\
DeepClusterV2 &
  96.87 &
  96.18 &
  \multicolumn{1}{c|}{96.25} &
  81.77 &
  82.47 &
  \multicolumn{1}{c|}{83.34} &
  83.59 &
  83.00 &
  \multicolumn{1}{c|}{83.20} &
  86.98 &
  86.53 &
  \multicolumn{1}{c|}{87.09} &
  83.13 &
  83.31 &
  82.79 \\
SwAV &
  94.53 &
  94.00 &
  \multicolumn{1}{c|}{94.10} &
  83.59 &
  83.55 &
  \multicolumn{1}{c|}{83.70} &
  82.56 &
  82.21 &
  \multicolumn{1}{c|}{83.09} &
  87.50 &
  87.12 &
  \multicolumn{1}{c|}{87.25} &
  82.87 &
  82.62 &
  83.39 \\ \hline
\end{tabular}
}
\end{table}

\begin{table}
  \centering
  \caption{Out of Domain Test. Linear Layer classifier with a non-frozen encoder trained on 350 images from the BBBC022 dataset and tested on the 200 images of the BBBC021 dataset.}
  \label{tab:generalisation}
  \setlength{\tabcolsep}{10pt} % Default value: 6pt
  \renewcommand{\arraystretch}{1.3}
  \begin{tabular}{cccc}
\hline
                & \multicolumn{3}{c}{\textbf{Linear Layer}} \\ \hline
\textbf{Method} & Acc             & F1             & AUC            \\ \hline
VGG 16  & 96.43          & 97.51      & 98.02        \\
ResNet18  & 91.52           & 93.21       & 92.87          \\
ConvNext  & \textbf{98.66}           &  \textbf{98.53}      &  \textbf{98.91}   \\
\hdashline
SimCLR          & 73.66           & 78.47          & 79.00          \\
Barlow Twins    & 54.91           & 62.02          & 58.76          \\
VICReg          & 37.95           & 40.00          & 39.21          \\
DeepClusterV2   & 72.32           & 75.99          & 76.12          \\
SwAV            & 56.25           & 57.50          & 57.30          \\ \hline
\end{tabular}
\end{table}

\section{Conclusion}
\label{sec:concl}
In this work, we conducted a thorough investigation to evaluate transfer and self-supervised representation learning on a large dataset in order to perform a downstream HCS quality control task. The quantitative results we obtained suggest that TL approaches perform better than SSL for this task.
Importantly, all SSL methods come with the need to choose crucial hyperparameters that will have significant impact on the learned representation. Among these hyperparameters are the choice of transformations that will define feature invariance in the obtained representation. Furthermore, SSL methods require an additional training on a large set of unannotated images. In contrast, an ImageNet pretrained encoder combined with a KNN downstream can be used out of the box and does not require any training or hyperparameter setting. If training can be performed, then unfreezing the encoder weights and fine tuning the training with a low amount of annotated data will slightly increase the performances, with TL still being a better option than SSL. Altogether this suggests that for the task of identifying abnormal versus normal image, transfer learning should be the preferred choice. 

Two reasons could be hypothesized to explain our findings. First, one could argue that our choice of transformations for the SSL approaches may not be the best option to create an optimal representation for our downstream quality control tasks. However, the choices we made were reasonable and relevant, and anyone seeking to solve a task using SSL would face the same issue: choosing hyperparameters and performing an additional training. Importantly, the debate on hyperparameter settings would be sound if transfer learning did not perform so well. Here we show that it is not only performing better than all SSL approaches, but it reaches almost perfect results in several setups, suggesting that even a better choice of SSL augmentations would not necessarily be worth finding. Secondly, this high performance obtained with transfer learning may be related to the specificity of the downstream task. Indeed, the experiments performed in the papers presenting these SSL approaches are often based on ImageNet classification which contains homogeneous semantic classes and therefore represents a different objective than the one presented in this work. Abnormal images do represent a very variable class with, for instance, out-of-focus image of cells being very different than an image containing debris. In this case, the low level features retrieved from the natural images of ImageNet may simply be sufficient and more efficient than higher semantic structure SSL representation typically provides. Although we focused on high-content screening here, we hope our findings will benefit quality control in other imaging modalities.

%\begin{comment}

\section{Acknowledgments}
\label{sec:acknowledgments}
This work was supported by ANR–10–LABX–54 MEMOLIFE and ANR–10 IDEX 0001
–02 PSL* Université Paris and was granted access
to the HPC resources of IDRIS under the allocation 2020-
AD011011495 made by GENCI.
%\end{comment}

\clearpage
\bibliographystyle{splncs04}
\bibliography{egbib}

\end{document}